# DeepJ: Graph Convolutional Transformers with Differentiable Pooling for Patient Trajectory Modeling

Deyi Li, MS[1], Zijun Yao, PhD[2], Muxuan Liang, PhD[3], Mei Liu, PhD[1]

[1]Department of Health outcomes and Biomedical Informatics, College of Medicine, University of Florida, Gainesville, Florida, USA

[2]Department of Electrical Engineering and Computer Science, University of Kansas, Lawrence, Kansas, USA

[3]Department of Biostatistics, University of Florida, Gainesville, Florida, USA

**Abstract**

*In recent years, graph learning has gained significant interest for modeling complex interactions among medical events in structured Electronic Health Record (EHR) data. However, existing graph-based approaches often work in a static manner, either restricting interactions within individual encounters or collapsing all historical encounters into a single snapshot. As a result, when it is necessary to identify meaningful groups of medical events spanning longitudinal encounters, existing methods are inadequate in modeling interactions cross encounters while accounting for temporal dependencies. To address this limitation, we introduce Deep Patient Journey (DeepJ), a novel graph convolutional transformer model with differentiable graph pooling to effectively capture intra-encounter and inter-encounter medical event interactions. DeepJ can identify groups of temporally and functionally related medical events, offering valuable insights into key event clusters pertinent to patient outcome prediction. DeepJ significantly outperformed five state-of-the-art baseline models while enhancing interpretability, demonstrating its potential for improved patient risk stratification.*

**Introduction**

The widespread availability of structured Electronic Health Record (EHR) data has enabled the development of accurate deep learning models for patient outcome prediction[1, 2]. While these deep learning models can effectively learn higher-order patient representations from structured EHR data, they often struggle to capture hidden relationships between medical events (or medical codes). These hidden relationships, which can be modeled as a graph structure, are especially important for personalized medicine, as they outline a patient's unique clinical trajectory and evolving health status while also capturing the complex medical codes interactions that shape this trajectory. Moreover, by encoding these hidden relationships as a graph and leveraging advanced graph machine learning models, we can enrich medical code representations by incorporating information from higher-order connectivity between medical codes[3]. By selecting an appropriate graph pooling method to aggregate these enriched codes into a single graph embedding vector representing the patient's entire trajectory, the predictive power of clinical outcome prediction models can be further enhanced, making this approach a promising avenue for improving predictive performance in clinical settings[4]. Lastly, compared to traditional population-level explainability approaches such as SHAP[5], this approach significantly enhances model interpretability, as it uncovers a unique clinical trajectory graph for each patient.

Recent efforts have attempted to uncover hidden graph structures from EHR data. Zhang et al.[6] and Choi et al.[7] constructed graphs using well-established ontology structures, such as the International Classification of Diseases (ICD) hierarchy. Choi et al.[8] introduced the Graph Convolutional Transformer (GCT), which leverages attention mechanisms to construct edges between medical codes. Moghaddam et al.[9] proposed Time-aware Personalized Graph Transformer (TPGT), further improving GCT's performance by incorporating temporal relationships across consecutive encounters.

However, these proposed models have several limitations. Firstly, models that rely on well-established ontology databases[6, 7] fail to capture relationships between different coding systems (e.g., RxNorm and ICD-10), as interactions across different coding systems are rarely supported. Moreover, the number of medical codes within a single encounter is often small, resulting in sparse node connections based on ontology relationship. Next, the GCT[8] and TPGT[9] models are limited to modeling graphical structures within a single encounter, overlooking longitudinal medical code interactions across multiple encounters. For instance, a diabetes diagnosis from a hospitalization one year ago may have led to an insulin prescription in a recent hospitalization, yet these models are unable to explicitly capture such dependencies. Lastly, all of these models use an embedding vector to represent the entire uncovered graph. However, the method used to derive this embedding vector suffers from insufficient utilization of information, reducing both prediction performance and interpretability. For example, in the GCT model, although the medical code sequence is modeled as a graph, the authors still use a single "visit token" for prediction[8], similar to the "classification token" used in BERT[10]. However, this representation may not always be the best global representation and significantly diminishes the model's interpretability. In the TPGT model, the authors use global mean pooling to derive an embedding vector for each encounter's medical code graph, followed by an LSTM with attention mechanism to aggregate these encounter-level graph embeddings for the final prediction[9]. However, this approach cannot separate the subgraph structures corresponding to distinct conditions, as patients may have multiple coexisting conditions. For example, a patient with both heart failure and lung cancer would have two distinct groups of medical codes spanning multiple encounters, each representing related comorbidities, lab tests, medications, and procedures.

To overcome these challenges, we propose **Dee**p **P**atient **J**ourney (DeepJ), a novel architecture designed to construct a personalized cross-encounter medical code interaction graph for each patient based on their entire encounter history. Our contributions in this study are as follows: (a) The **G**raph **S**tructure **L**earning (GSL) component in DeepJ extended the GCT architecture[8] to captures both intra-encounter and inter-encounter code connectivity. (b) The **C**linical **M**odule **D**iscovery (CMD) component in DeepJ, a hierarchical clustering module adapted from the differential graph pooling method[11], is able to identify clinically meaningful subgroups – clusters of codes that are closely related and interact with each other longitudinally across encounters. (c) Through comprehensive evaluation on two datasets – the public eICU dataset[12] and the private University of Kansas Medical Center (KUMC) inpatient dataset – across two clinical prediction tasks: ICU mortality prediction and Acute Kidney Injury (AKI) onset prediction, DeepJ significantly outperformed five state-of-the-art baseline models while demonstrating strong interpretability.

## Methods

### Problem Formulation

To enhance clarity, we illustrate our proposed methodology using a single patient as an example. The EHR dataset consists of a vocabulary of medical codes denoted as $c_1, c_2, \ldots, c_{|C|} \in C$, with a vocabulary size of $|C|$. A patient's clinical trajectory is modeled as a sequence of temporally ordered encounters, denoted as $V_1, V_2, \ldots, V_p \ldots, V_P$, where $P$ denotes the total number of encounters and each encounter $V_p$ contains a subset of unique medical codes $V_p \in C$. The same codes may appear across different encounters of the same patient. The elapsed time from the current encounter to the first encounter is denoted as $t_1, t_2, \ldots, t_p, \ldots, t_P$, where $t_1 = 0$.

Given $P_{max}$ as the maximum number of encounters a patient can have in their history, $C_{max}$ as the maximum number of medical codes in an encounter, we represent a patient as $V = \left[c_1^{(1)}, c_2^{(1)}, \ldots, c_{C_{max}}^{(1)}, c_1^{(2)}, c_2^{(2)}, \ldots, c_{C_{max}}^{(2)}, \ldots, c_{C_{max}}^{(P_{max})}\right]$ and the elapsed time $t_1, t_2, \ldots, t_{P_{max}}$. The tasks are formulated as follows: (1) uncovering a hidden intra-encounter and inter-encounter graph; (2) clustering the uncovered graph into clinical modules to identify meaningful subgroups of co-occurring medical codes across encounters; and (3) predicting binary outcomes (e.g., ICU mortality and AKI onset) using the uncovered patient-specific graph.

### Input Embedding and Time Encoding

The architecture of DeepJ is presented in **Figure 1**. Each code in the input sequence is projected to a vector representation by a learnable embedding matrix , where $d_{model}$ denotes the embedding size. Consequently, the entire input sequence is transformed into an embedding matrix $E \in \mathbb{R}^{SeqLen \times d_{model}}$, where $SeqLen = P_{max} C_{max}$. To capture temporal relationships between encounters, time encoding (TE)[13] is used:

$$TE(t_p, d) = \begin{cases} \sin(t_p / t_{max}^{\frac{2d}{d_{model}}}), & d \text{ is even} \\ \cos(t_p / t_{max}^{\frac{2d}{d_{model}}}), & d \text{ is odd} \end{cases}$$

Here, $t_{max}$ denotes the maximum elapsed time in the dataset, and $d$ denotes index of the features. All codes within an encounter share the same timestamp. The input to the first block of the model is computed as $Z = E + TE(V) \in \mathbb{R}^{SeqLen \times d_{model}}$.

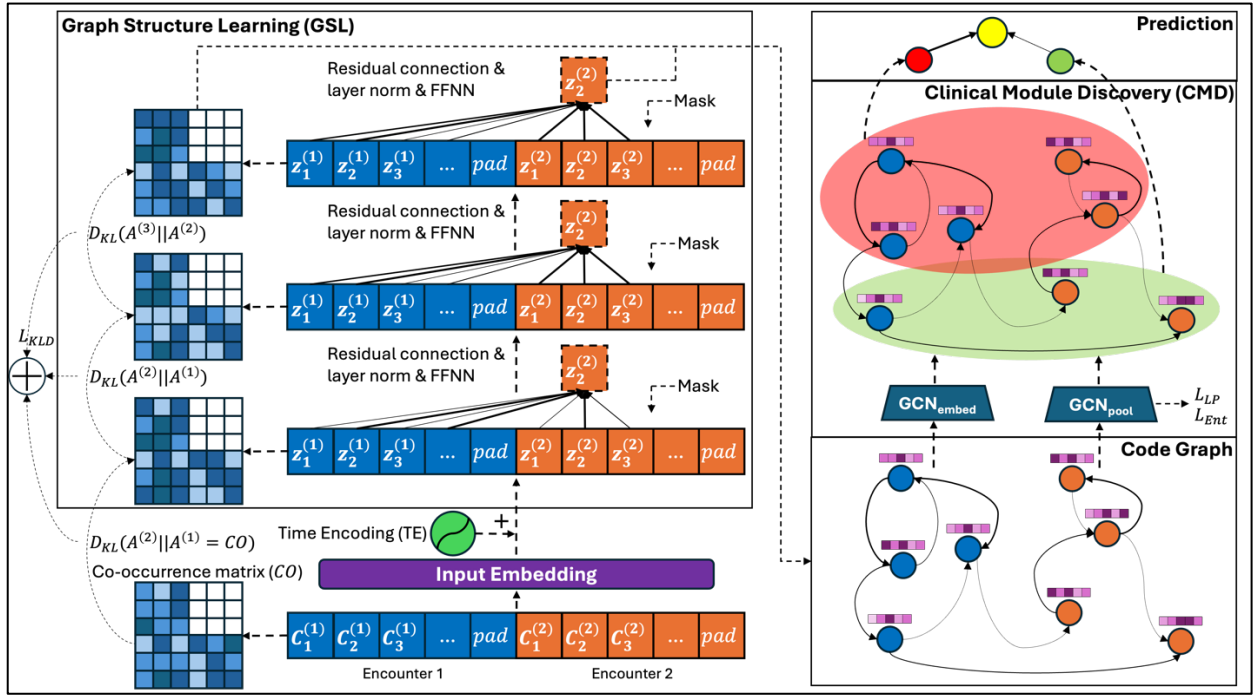

**Figure 1:** The architecture of DeepJ. The model consists of two key components: Graph Structure Learning (GSL) and Clinical Module Discovery (CMD). FFNN, feedforward neural network. GCN, graph convolutional network.

**Graph Structure Learning**

A transformer-based architecture is adopted to learn hidden intra-encounter and inter-encounter graph structures. For each code embedding $z_i$ in $Z$, a masked self-attention mechanism is used to learn edges connecting code $j$ to code $i$, where code $j$ may belong to either a previous encounter or the same encounter as code $i$. Specifically, each $z_i$ in $Z$ is projected into a key vector, a query vector, and a value vector, $k, q, v \in \mathbb{R}^{d_{model}}$, using three learnable matrices $W_k, W_q, W_v \in \mathbb{R}^{d_{model} \times d_{model}}$, respectively. The attention scores are computed via scaled dot-product attention:

$$\alpha_{i,j} = \frac{q_i \cdot k_j^T}{\sqrt{d_{model}}} + \mathcal{M}_{i,j}$$

Here, $\alpha_{i,j}$ represents the attention score between code $i$ and code $j$. $\mathcal{M}_{i,j}$ is a masking term defined as follows:

$$\mathcal{M}_{i,j} = \begin{cases} -\infty, & \text{code } i \text{ happend before code } j; \text{ either code } i \text{ or } j \text{ is a padding code} \\ 0, & \text{otherwise} \end{cases}$$

This mask ensures that (1) future codes do not influence past codes. In other words, in the learned graph, an edge pointing from a future code to a past code will not be allowed. (2) padding codes do not receive any attention, preventing them from contributing to the learned representations. The attention weights are obtained via softmax:

$$w_{i,j} = \frac{\exp(\alpha_{i,j})}{\sum_{j'} \exp(\alpha_{i,j'})}$$

Here, $w_{i,j}$ denotes the attention weight between code $i$ and code $j$. Finally, the updated embedding for each code $i$, denoted as $\tilde{e}_i$, is computed as:

$$\tilde{e}_i = \sum_j w_{i,j} v_j \in \mathbb{R}^{d_{model}}$$

The updated encounter sequence embedding $\tilde{E} \in \mathbb{R}^{SeqLen \times d_{model}}$, is then passed through a feed-forward neural network (FFNN) to capture non-linear interactions. The masked self-attention layer and the FFNN, along with residual connections and layer normalization, together constitute an **E**xtended **G**raph **C**onvolutional **T**ransformer (EGCT) block. Stacking $N$ EGCT blocks together forms the **G**raph **S**tructure **L**earning (GSL) component. GSL learns the higher-order representation for each code, and the attention weights from the last EGCT block ($A^{(N)}$) serve as the adjacency matrix $\mathcal{A}$ of the learned graph.

Inspired by Choi et al.[8], the attention weight matrix of the first transformer block is replaced by a co-occurrence matrix, which captures the real-world statistical relationships between codes, calculated as follows:

$$CO_{i,j} = P\left(c_i^{(p_2)} \middle| c_j^{(p_1)}\right) = \frac{P(c_j^{(p_1)}, c_i^{(p_2)})}{P(c_j^{(p_1)})}, \text{where } p_1 \leq p_2$$

To ensure continuity, the attention weight matrix of each block is encouraged to remain similar to that of the preceding block, an auxiliary loss term is introduced:

$$L_{KLD} = \sum_{n=2}^{N} L_{KLD}^{(n)} = \sum_{n=2}^{N} D_{KL}(A^{(n-1)} || A^{(n)}), \text{where } A^{(1)} = CO$$

Here, $N$ denotes the number of EGCT blocks and $D_{KL}$ denotes the Kullback-Leibler (KL) divergence. This design ensures that the learned edge weights align with real-world conditions while allowing flexibility to learn new edges.

**Clinical Module Discovery**

The output of the last EGCT block of GSL consists of higher-order representation $\tilde{e}_i^{(N)}$ of each code, forming node representations $X$ when viewed as nodes in a graph (i.e., $X = \tilde{E}^{(N)}$), and an adjacency matrix $\mathcal{A}$ generated from the last attention weight matrix (i.e., $\mathcal{A} = A^{(N)}$), which describes one patient's personalized medical code graph connections and edge weights. They serve as the input for the **C**linical **M**odule **D**iscovery (CMD).

CMD consists of **diff**erential **pool**ing (DiffPool) blocks[11] with each DiffPool block aiming to cluster the input graph into several subgraphs, facilitating the discovery of cross-encounter, interacting cluster of medical events. We stacked $M$ DiffPool blocks together to enable hierarchical graph pooling, resulting in multi-level graph resolutions. In the $m$-th DiffPool block, two graph convolutional network (GCN) layers[14] were employed in parallel—one for aggregating neighborhood information to enhance node representations and the other for projecting the nodes into $g^{m+1}$ clusters, where $g^{m+1}$ is pre-defined by the user:

$$H^{(m)} = GCN_{m,embed}(\mathcal{A}^{(m)}, X^{(m)}) \in \mathbb{R}^{g^{(m)} \times d_{model}}$$

$$S^{(m)} = softmax(GCN_{m,pool}(\mathcal{A}^{(m)}, X^{(m)})) \in \mathbb{R}^{g^{(m)} \times g^{(m+1)}}$$

Here, $H^{(m)}$ denotes the updated node representations obtained through GCN message passing and $S^{(m)}$ denotes the soft cluster assignments. We further added residual connections and batch normalization to facilitate training stability. Then, the node representations and adjacency matrix of the pooled graph is derived by matrix multiplication:

$$X^{(m+1)} = S^{(m)^T} H^{(m)} \in \mathbb{R}^{g^{(m+1)} \times d_{model}}$$

$$\mathcal{A}^{(m+1)} = S^{(m)^T} \mathcal{A}^{(m)} S^{(m)} \in \mathbb{R}^{g^{(m+1)} \times g^{(m+1)}}$$

For each DiffPool block, two auxiliary loss terms, link prediction loss and entropy regularization loss, were computed to facilitate training. Link prediction loss $L_{LP}$ encourages the node similarity matrix recovered by the node assignment matrix $S^{(m)}$ to be similar to the input adjacency matrix $\mathcal{A}^{(m)}$:

$$L_{LP} = \sum_{m=1}^{M} L_{LP}^{(m)} = \sum_{m=1}^{M} \left\| \mathcal{A}^{(m)} - S^{(m)} S^{(m)^T} \right\|_F$$

Here, $\|\cdot\|_F$ denotes the Frobenius norm. Entropy regularization loss encourages the cluster assignment for each node to be "hard", promoting more unambiguous cluster memberships:

$$L_{Ent} = \sum_{m=1}^{M} L_{Ent}^{(m)} = \sum_{m=1}^{M} \sum_{r=1}^{g^{(m)}} Ent(S_r^{(m)})$$

Here, $Ent$ denotes the entropy function and $r$ denotes the $r$-th row of the $S^{(m)}$ matrix.

**Clinical Module Weighting and Prediction**

In the last DiffPool block of CMD, each node of the pooled graph represents a group of nodes from the original graph. In the **C**linical **M**odule **W**eighting (CMW), a simple attention mechanism is employed to compute a weighted sum of the pooled graph node representations :

$$\alpha_q = softmax(X_q^{(M+1)} w^T)$$

$$G_{final} = \sum_{q=1}^{g^M} \alpha_q X_q^{(M+1)} \in \mathbb{R}^{d_{model}}$$

Here, $q$ indexes the $q$-th cluster embedding produced by the last DiffPool block. $w \in \mathbb{R}^{d_{model}}$ is a learnable parameter vector. $G_{final}$ denotes the final vector that embeds the entire encounter sequence of the patient. Finally, $G_{final}$ is fed into a FFNN classifier for outcome prediction. The final loss of the model is calculated as follows:

$$L_{model} = L_{predict} + \lambda_{KLD} L_{KLD} + \lambda_{LP} L_{LP} + \lambda_{Ent} L_{Ent}$$

Here, $L_{predict}$ represents the loss for the outcome prediction task. In this study, we used the negative log-likelihood loss for $L_{predict}$. $\lambda_{KLD}$, $\lambda_{LP}$ and $\lambda_{Ent}$ control the trade-offs between the KL-divergence loss from GSL, the link prediction loss, and the entropy regularization loss from CMD, respectively.

**Data Source and Processing**

Two datasets were used to evaluate the performance of DeepJ: the eICU dataset[12] and the University of Kansas Medical Center (KUMC) Inpatient dataset. The eICU dataset is a publicly available dataset containing records of patients admitted to critical care units in 2014 and 2015. Data processing followed the same pipeline as in Choi et al[8]. Specifically, two types of medical code were collected: diagnoses and treatments. ICU encounters longer than 24 hours were excluded. We considered the most recent two ICU encounters. The prediction task on the eICU dataset was to predict ICU mortality for the most recent ICU encounter.

The KUMC Inpatient dataset consists of hospitalization records of patients from January 1, 2016, to December 31, 2020. The prediction task for this dataset was the onset of AKI, defined according to the KDIGO clinical practice guidelines[15]. Similar to the eICU dataset, each encounter was defined as a 24-hour observation window between admission and the day of AKI onset. For patients who did not develop AKI, the counterpart to the AKI onset day was set as the last recorded serum creatinine (SCr) measurement[16]. If the time between admission and AKI onset or the last SCr measurement exceeds three days, we randomly sampled three encounters within this window for each patient. Finally, for each encounter, two types of medical codes were collected: diagnoses and medications.

**Experimental Settings and Analysis**

We evaluated the performance of DeepJ against five baseline models: GCN[14], LSTM[17], Deepr[18], Transformer[19], and GCT[8]. For GCN, the code co-occurrence matrix $CO$ was used as the adjacency matrix and code embeddings was used as node features. For Transformer and GCT, to ensure a fair comparison, we applied the same masking mechanism as DeepJ to prevent future codes from contributing information to past codes. Additionally, we controlled for model complexity by setting the number of blocks (or layers) in GCN, Transformer, GCT, and DeepJ to ensure that they had approximately the same number of parameters. To address the over-smoothing problem in deep GCN models[20], we incorporated residual connections and batch normalization in GCN. In DeepJ, the coefficients of the three auxiliary loss terms ($\lambda_{KLD}$, $\lambda_{LP}$ and $\lambda_{Ent}$) were all set to 1.0. We conducted two sets of ablation experiments to evaluate the impact of two key components, GSL and CMD, on DeepJ's performance. Specifically, in the DeepJ w/o GSL model, $CO$ was enforced across all EGCT blocks, preventing the model from learning any new edges. In the DeepJ w/o CMD model, the output embeddings of GSL were mean-pooled and directly passed through the classifier, bypassing CMD. All models were evaluated using 10-fold cross-validation on both datasets. We reported the mean and 95% confidence interval (CI) for four performance metrics: recall, F1-score, AUROC, and AUPRC.

To interpret the GSL module, we analyzed intra-encounter and inter-encounter edge statistics in the eICU dataset, focusing on edge frequency and weights. For the CMD module, we analyzed the top five medical codes most frequently grouped into the same clinical module as four common ICU conditions—acute renal failure, hypertension, hypotension, and hyperkalemia—in the eICU dataset. This method allows us to examine the interpretability of DeepJ at the population level, rather than limiting evaluation to specific patient cases with good interpretability.

**Results**

The statistics of the two datasets are summarized in **Table 1**. The eICU dataset contains 39,874 patients, while the KUMC inpatient dataset consists of 25,955 patients. The prediction task for the eICU dataset was ICU mortality, with a positive class ratio of 7.48%. The prediction task for the KUMC Inpatient dataset was AKI onset, with a positive class ratio of 13.36%.

**Table 1:** Statistical information of the datasets.

|  | eICU | KUMC |
|---|---|---|
| Number of patients | 39,874 | 25,955 |
| Prediction Task | Mortality | AKI onset |
| Positive case ratio (%) | 7.48 | 13.36 |
| Window size of an encounter (hours) | 24 | 24 |
| Average number of codes in an encounter | 11.48 | 34.46 |
| Max elapsed time from admission (days) | 54.49 | 40.00 |

The performance of the baseline models and the results of the DeepJ ablation experiments are presented in **Table 2**. On the eICU dataset for the ICU mortality prediction task, the full DeepJ model outperformed all baseline models on three out of four metrics, except for recall. In the ablation experiments, both the $CO$-enforced model (DeepJ w/o GSL) and the no graph pooling model (DeepJ w/o CMD) showed significantly lower performance compared to the full DeepJ model. Similar trends were observed on the KUMC dataset for the AKI prediction task, where the full DeepJ model achieved the highest performance across all four metrics, further demonstrating the importance of GSL and CMD modules in improving prediction performance.

**Table 2:** Model performance on the eICU and KUMC datasets. The numbers in parentheses represent the 95% CI.

|  | eICU | | | | KUMC | | | |
|---|---|---|---|---|---|---|---|---|
|  | **Recall** | **F1** | **AUROC** | **AUPRC** | **Recall** | **F1** | **AUROC** | **AUPRC** |
| **GCN** | **0.7250 (0.7094 - 0.7406)** | 0.7542 (0.7422 - 0.7663) | 0.8737 (0.8642 - 0.8832) | 0.5552 (0.5280 - 0.5824) | 0.5346 (0.5199 - 0.5492) | 0.5335 (0.5144 - 0.5526) | 0.6379 (0.6285 - 0.6473) | 0.2162 (0.2081 - 0.2244) |
| **LSTM** | 0.5000 (0.5000 - 0.5000) | 0.4806 (0.4805 - 0.4806) | 0.7609 (0.7108 - 0.8110) | 0.2711 (0.1703 - 0.3719) | 0.5001 (0.4998 - 0.5005) | 0.4645 (0.4638 - 0.4652) | 0.6714 (0.6636 - 0.6791) | 0.2700 (0.2575 - 0.2826) |
| **Deepr** | 0.7216 (0.7084 - 0.7349) | 0.7559 (0.7451 - 0.7667) | 0.8878 (0.8808 - 0.8948) | 0.5661 (0.5402 - 0.5920) | 0.5440 (0.5398 - 0.5481) | 0.5505 (0.5447 - 0.5562) | 0.6625 (0.6520 - 0.6731) | 0.2403 (0.2321 - 0.2484) |
| **Transformer** | 0.6379 (0.6099 - 0.6658) | 0.6833 (0.6570 - 0.7097) | 0.8655 (0.8574 - 0.8736) | 0.5133 (0.4874 - 0.5391) | 0.5032 (0.5002 - 0.5061) | 0.4720 (0.4654 - 0.4785) | 0.6702 (0.6596 - 0.6808) | 0.2481 (0.2312 - 0.2650) |
| **GCT** | 0.6501 (0.6223 - 0.6778) | 0.6991 (0.6727 - 0.7255) | 0.8746 (0.8682 - 0.8810) | 0.5342 (0.5127 - 0.5557) | 0.5063 (0.5041 - 0.5084) | 0.4803 (0.4754 - 0.4852) | 0.6745 (0.6653 - 0.6836) | 0.2595 (0.2474 - 0.2716) |
| **DeepJ w/o GSL** | 0.7133 (0.7057 - 0.7209) | 0.7612 (0.7537 - 0.7687) | 0.8864 (0.8783 - 0.8944) | 0.5777 (0.5557 - 0.5996) | 0.5494 (0.5419 - 0.5570) | 0.5580 (0.5464 - 0.5695) | 0.6790 (0.6698 - 0.6882) | 0.2871 (0.2751 - 0.2992) |
| **DeepJ w/o CMD** | 0.6312 (0.6052 - 0.6573) | 0.6806 (0.6478 - 0.7134) | 0.8747 (0.8673 - 0.8821) | 0.5422 (0.5230 - 0.5613) | 0.5073 (0.5021 - 0.5125) | 0.4812 (0.4708 - 0.4916) | 0.6781 (0.6670 - 0.6891) | 0.2766 (0.2623 - 0.2909) |
| **DeepJ** | 0.7228 (0.7161 - 0.7294) | **0.7684 (0.7608 - 0.7760)** | **0.8933 (0.8836 - 0.9030)** | **0.5906 (0.5624 - 0.6189)** | **0.5566 (0.5412 - 0.5720)** | **0.5667 (0.5423 - 0.5911)** | **0.7265 (0.7149 - 0.7381)** | **0.3317 (0.3183 - 0.3451)** |

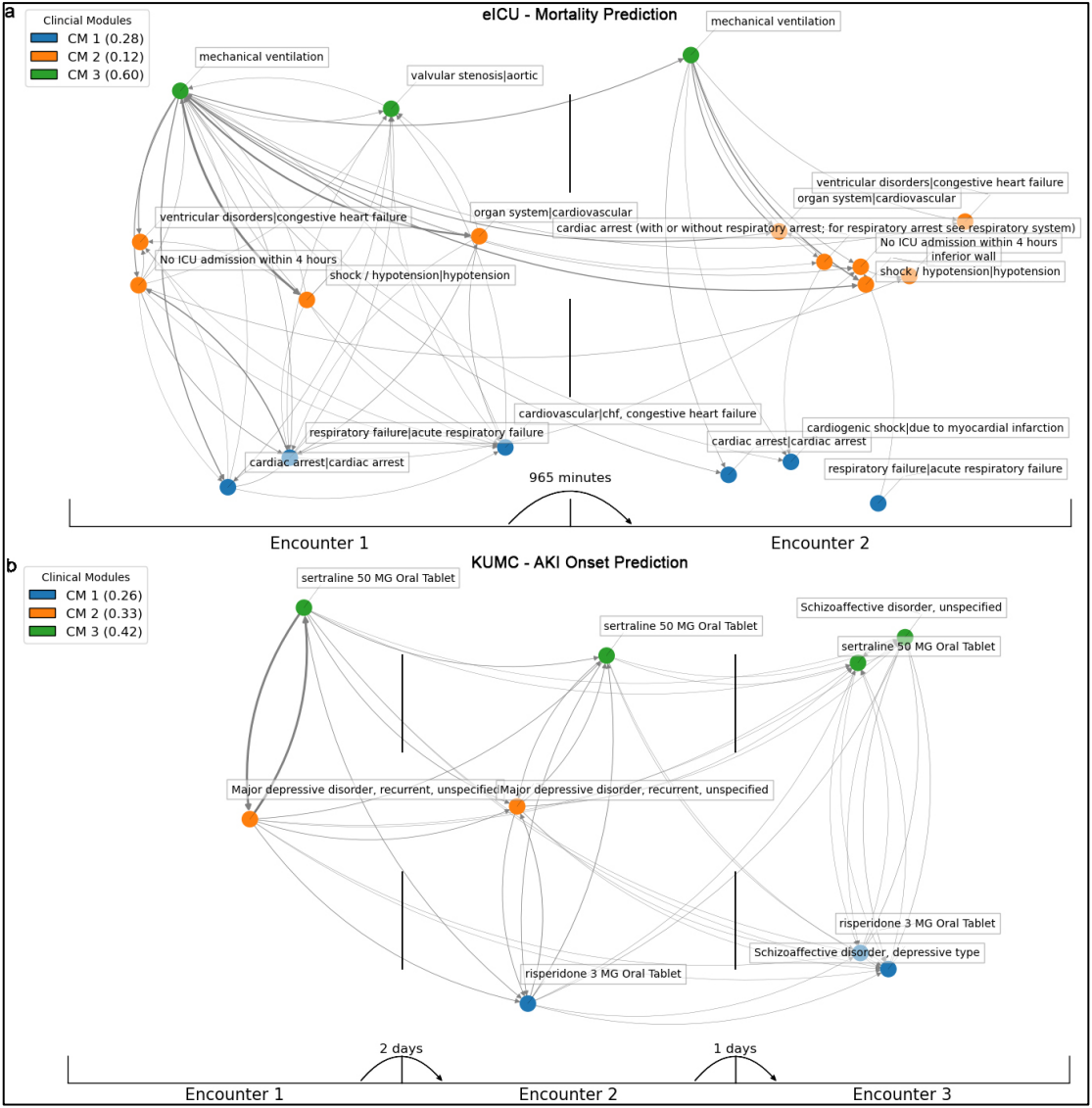

**Figure 2:** Uncovered patient trajectory graph structures of two example patients. Nodes are clustered into different clinical modules (CM), and the weight of each clinical module is shown in the legend. Edge weights are indicated by edge boldness. Edges with a weight below 0.1 are filtered out. **a** eICU dataset. **b** KUMC dataset.

Two examples of uncovered hidden graphs, one from each dataset, are presented in **Figure 2**. In the first example (eICU dataset, **Figure 2 a**), the patient had two ICU admissions, with an interval of 965 minutes between them. During the first ICU stay, the most salient edges were: (1) mechanical ventilation → hypotension, (2) mechanical ventilation → cardiovascular symptoms, (3) mechanical ventilation → cardiac arrest, (4) mechanical ventilation → congestive heart failure, (5) cardiac arrest → no ICU admission within 4 hours.

During the second ICU stay, the most salient edges were: (1) ventilation → hypotension, (2) mechanical ventilation → cardiovascular symptoms. The most salient inter-encounter edge was mechanical ventilation → hypotension. In this case, DeepJ identified three clinical modules: (1) the blue module was primarily associated with mechanical ventilation; (2) the orange module was dominated by cardiovascular symptoms; and (3) the green module was a mixture of respiratory and cardiovascular symptoms. The attention score indicated that the green module (mechanical ventilation) was the most important in predicting ICU mortality (score = 0.60).

In the second example (KUMC dataset, **Figure 2 b**), the patient had three encounters, with intervals of 2 days and 1 day between them, respectively. The most salient edges appeared in the first encounter, highlighting a mutual connection between sertraline and major depressive disorder. Some of the inter-encounter edges included: sertraline → sertraline, sertraline → schizoaffective disorder, risperidone → schizoaffective disorder. DeepJ identified the sertraline module (green) as the most important for AKI onset prediction (score = 0.42).

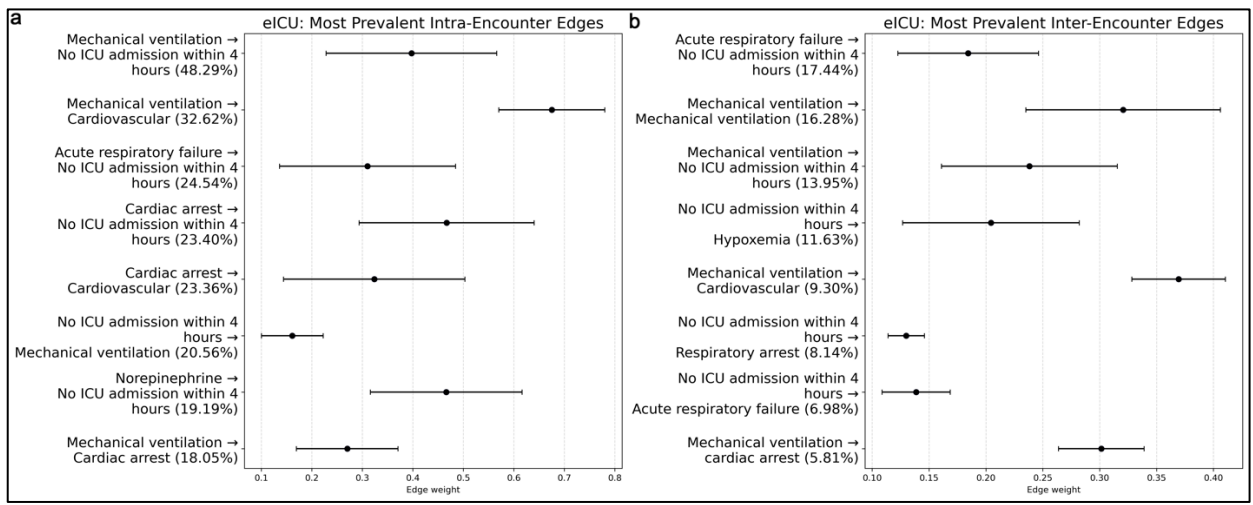

**Figure 3:** Top eight most prevalent intra-encounter and inter-encounter edges in the eICU dataset. The prevalence of each edge is marked on the y-axis labels. The dot and error bar represent the mean edge weights and standard deviation. **a** Intra-encounter edges. **b** Inter-encounter edges.

The edge statistics of the eICU dataset are presented in **Figure 3**, highlighting key intra- and inter-encounter connections. Some of the most frequent intra-encounter edges included the mutual connections between no ICU admission within 4 hours and mechanical ventilation (48.29% and 20.56%), as well as mechanical ventilation → cardiovascular symptoms (32.62%). Notably, mechanical ventilation → cardiovascular symptoms also had the highest edge weight, with a mean value of 0.68. Among the most frequent inter-encounter edge connections were acute respiratory failure → mechanical ventilation and mechanical ventilation → mechanical ventilation. Additionally, mechanical ventilation → cardiovascular symptoms was the most important inter-encounter edge, with a mean edge weight of 0.37.

**Figure 4** illustrates the codes most frequently grouped together in the same clinical modules as four common ICU conditions: acute renal failure, hypertension, hypotension, and hyperkalemia. For acute renal failure, some of the top co-clustered codes included acute respiratory failure, chest X-ray, and sepsis. For hypertension, the top co-clustered codes included aspirin, hyperlipidemia, and no recent ICU admission. The results for hypotension and hyperkalemia are presented in **Figure 4 c** and **d**.

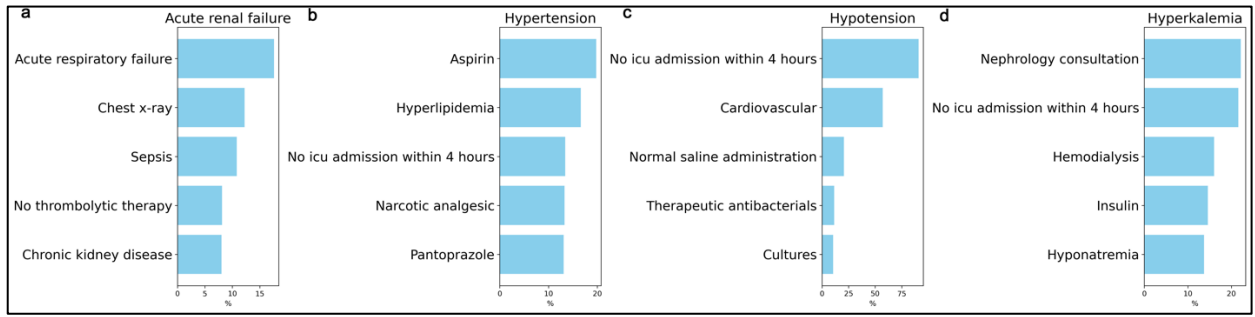

**Figure 4:** Top five codes that are most frequently grouped into the same clinical module as the four common ICU symptoms: **a** Acute renal failure. **b** Hypertension. **c** Hypotension. **d** Hyperkalemia.

**Discussion and Conclusions**

In this study, we propose a novel architecture, DeepJ, which combines Transformer and graph-based models to learn hidden graph structures between medical codes in EHR data. Compared to previous research, our contributions are the following: (1) ***Modeling Cross-Encounter Relationship***: we extended the GCT architecture from Choi et al.[8] to handle multiple encounters by introducing an additional Temporal Encoding (TE) module and a causal mask, enabling the model to capture dependencies across encounters with arbitrary time spans. (2) ***Clinical Module Discovery (CMD)***: we introduced the CMD module, composed of DiffPool blocks, allowing the model to discover interacting medical codes spanning multiple encounters, forming meaningful clinical modules. (3) ***Enhanced Performance***: leveraging high-quality graph structures and graph pooling, DeepJ outperformed all investigated baseline models on two different clinical prediction tasks across two datasets.

Our results, shown in **Table 2**, reveal that enforcing the co-occurrence matrix ***CO*** across all EGCT blocks (DeepJ w/o GSL) results in lower performance compared to enforcing it only in the first EGCT block (DeepJ full model). This suggests that allowing the GSL module with flexibility to learn new edges is beneficial. When the CMD module was removed, there was a significantly decline in predictive performance, indicating the effectiveness of dividing the graph into subgraphs and applying attention-weighted pooling. This suggests that not all clinical modules contribute equally to outcome prediction. For instance, in **Figure 2 a**, the cardiovascular-respiratory mixed module (blue, 0.28) was more important than the cardiovascular-only module (orange, 0.12), suggesting that multi-organ acute conditions played a more critical role in predicting patient mortality.

As **Figure 3** illustrates, mechanical ventilation and no ICU admission within 4 hours were the most frequently occurring nodes, as they are among the most common codes in eICU encounters and strong predictors of mortality, with frequent ICU admissions indicating a higher risk for mortality. Beyond these two codes, several meaningful and well-evidenced edges were captured, such as cardiac arrest → cardiovascular symptoms and mechanical ventilation → cardiac arrest, indicating the effectiveness of the GSL module[21]. In **Figure 4**, conditions such as acute respiratory failure, sepsis, and chronic renal disease were frequently grouped into the same clinical module as acute renal failure, and aspirin and hyperlipidemia were frequently grouped with hypertension, further validating the CMD module's capability to capture well-documented clinical associations[22].

A limitation of this study is that the uncovered graphs represent associations rather than causal relationships. Some edge connections may not accurately reflect real-world scenarios. Further research should focus on enhancing causal inference methods to improve the interpretability and reliability of the learned graph structures. In conclusion, we introduce DeepJ, a novel model for learning graph structures from EHR data, laying the foundation for future research in graph learning for EHR analysis.

**Code Availability**

The code of this study can be found at https://github.com/GatorAIM/DeePJ.

**Acknowledgement**

This project is supported by the NIH/NIDDK under award number R01DK137881.